\documentclass[runningheads]{llncs}

\usepackage[mobile]{eccv}

\usepackage{eccvabbrv}

\usepackage{graphicx}
\usepackage{booktabs}
\usepackage{amsmath}
\usepackage{amssymb}
\usepackage{booktabs}
\usepackage{mathtools}
\usepackage{multirow}
\usepackage{mismath}
\usepackage{eqnarray}
\usepackage{colortbl}
\usepackage{ragged2e}
\usepackage{tabularray}

\usepackage[accsupp]{axessibility}  

\usepackage{hyperref}

\usepackage{orcidlink}
\definecolor{frenchblue}{rgb}{0.27, 0.45, 0.77}
\definecolor{kellygreen}{rgb}{0.44, 0.68, 0.28}

\begin{document}

\title{IG-FIQA: Improving Face Image Quality Assessment through Intra-class Variance Guidance robust to Inaccurate Pseudo-Labels} 

\titlerunning{Abbreviated paper title}

\author{
Minsoo Kim\inst{1,2}\orcidlink{0000-0001-6813-7335} \and
Gi Pyo Nam\inst{1,2}\orcidlink{0000-0002-3383-7806} \and
Haksub Kim\inst{1}\orcidlink{0000-0002-8780-9747} \and
Haesol Park\inst{1}\orcidlink{0009-0002-7615-6231} \and
Ig-Jae Kim\inst{1,2,3}\orcidlink{0000-0002-2741-7047}
}

\authorrunning{M.~Kim et al.}

\institute{Korea Institute of Sciene and Technology, Korea \and
Korea National University of Science and Technology, Korea \and
Yonsei-KIST Convergence Research Institute, Yonsei University\\
\email{\{kim1102, gpnam, hskim, haesol, drjay\}@kist.re.kr}}

\maketitle

\begin{abstract}
In the realm of face image quality assesment (FIQA), method based on sample relative classification have shown impressive performance. However, the quality scores used as pseudo-labels assigned from images of classes with low intra-class variance could be unrelated to the actual quality in this method. To address this issue, we present IG-FIQA, a novel approach to guide FIQA training, introducing a weight parameter to alleviate the adverse impact of these classes. This method involves estimating sample intra-class variance at each iteration during training, ensuring minimal computational overhead and straightforward implementation. Furthermore, this paper proposes an on-the-fly data augmentation methodology for improved generalization performance in FIQA. On various benchmark datasets, our proposed method, \textbf{IG-FIQA}, achieved novel state-of-the-art (SOTA) performance.


\end{abstract}

\section{Introduction}

\label{sec:intro}

\begin{figure*}[t]
    \begin{subfigure}[ht]{0.45\textwidth}
    \centering
	\includegraphics[width=1.0\columnwidth]{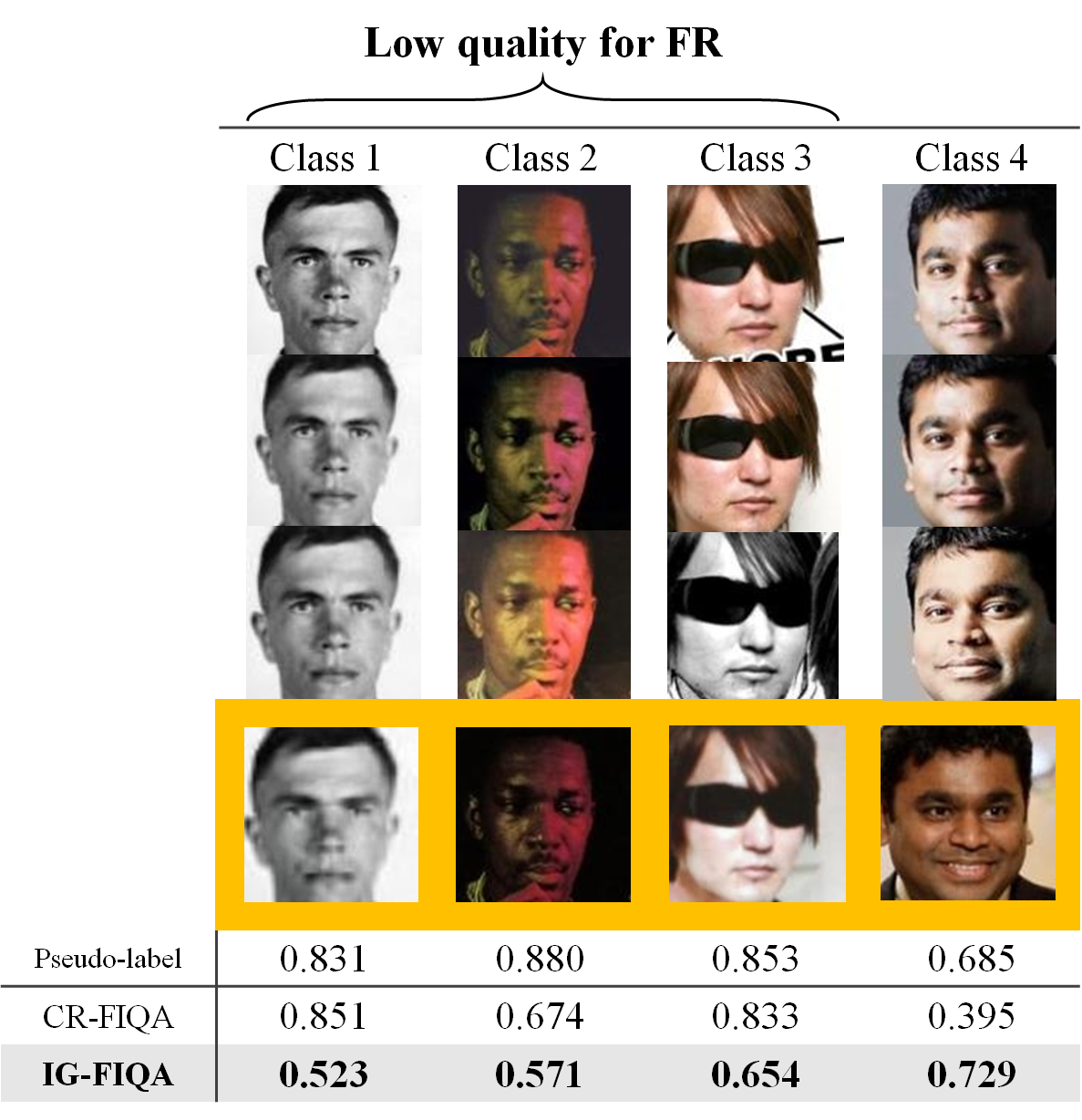}
	\caption{
		Example of mislabeled pseudo label that generated using conventional SOTA \cite{boutros2023cr} in the MS1M-V2 dataset. The table lists the min-max normalized scores measured by various methods on images with yellow boxed.
	}
	\label{fig1}
    \end{subfigure}
    \hfill
    \begin{subfigure}[ht]{0.53\textwidth}
    \centering
       \includegraphics[width=1.0\linewidth]{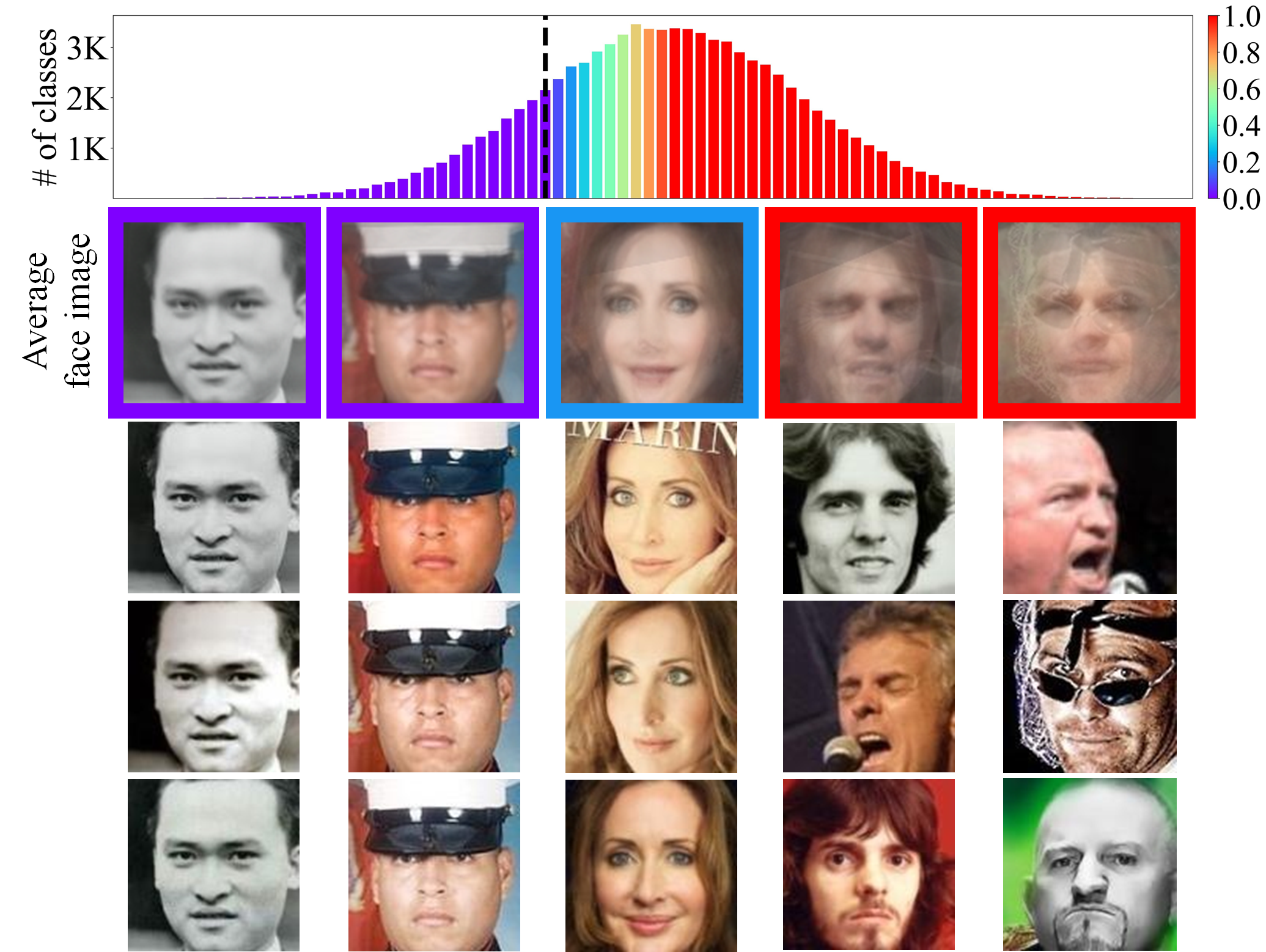}
	\caption{
		Illustration of the distribution of the proposed weights used for loss calculation in the MS1M-V2 dataset. To create an average face image, we randomly sampled 20 images from each class.
	}
	\label{weight_average}
    \end{subfigure}
    \caption{The mislabeling problem existing in conventional SOTA method and our suggesting solution. Fig. \ref{weight_average} depicts our proposed method, which can ignore classes with low intra-class variance (colored purple) during training, while classes with high intra-class variance (colored red) are fully utilized for training the FIQA regression network.}
    \label{w_distrib_fig}
\end{figure*}

%
Facial Image Quality Assessment (FIQA) aims to estimate the quality of facial images for ensuring the reliability of face recognition (FR) algorithms \cite{grother2014face}. Unlike traditional image quality assessment methods~\cite{mittal2012no, liu2017rankiqa, wang2004image, bosse2017deep, zhang2011fsim, mittal2012making}, which focus on the low-level image characteristics such as brightness, distortion, and sharpness, FIQA also considers the factors that affect the FR performance, such as pose variation, facial expression, and occlusion. For example, a high-resolution facial image with a face mask can receive a high-quality score in image quality assessment (IQA). However, the same image could get a lower score in FIQA because the mask interferes with FR.

Recent methods in FIQA can be categorized into two types: methods that propose computational measurements by analyzing the pre-trained FR feature space, and methods that predict the FIQ by training regression networks. Among them, the regression-based methods focus on generating appropriate pseudo labels to train the FIQA regression network consistently. Various approaches have been proposed for this purpose, such as manual labeling \cite{best2018learning}, Wasserstein Distance (WD) \cite{ou2021sdd}, and Certainty Ratio (CR) \cite{boutros2023cr}. Regression models trained using CR achieve state-of-the-art performance on various benchmarks, demonstrating the effectiveness of using sample relative classifiability as an approximation for face image quality. CR is computed by combining the similarity between the embedding feature and the positive class centroid ($cos(\theta_{y_{i}})$) with the similarity between the embedding feature and the nearest negative class centroid $(cos(\theta_{y_{j,j\neq i}}))$. It is designed to have a higher value when the similarity $cos(\theta_{y_{i}})$ is closer and the similarity $cos(\theta_{y_{j,j\neq i}})$ is further away.


FIQA method leveraging sample relative classifiability have achieved remarkable performance but still have limitations. The first limitation is that pseudo-labels generated from classes with low intra-class variance cannot accurately reflect the quality of the samples. 
Typically, when determining pseudo-labels for image quality, the similarity between the embedding and the centroid of the corresponding class is utilized. However, in cases where intra-class variance is low, meaning of consist similar images, a high similarity is calculated, leading to the generation of incorrect pseudo-labels regardless of the actual image quality. As seen in Fig. \ref{fig1}, even low-resolution (column 1), low-light (column 2), and occluded (column 3) face images are assigned higher pseudo-label than high-quality face images (column 4) due to low intra-class variance.
This is a common problem because real-world datasets are often collected from the web~\cite{guo2016ms, cao2018celeb, cao2018vggface2, wang2018devil, parkhi2015deep}, and the removal of noisy data relies on automated methods that use the feature similarities~\cite{deng2020sub, zhu2021webface260m, deng2017marginal}. As a result, classes with low intra-class variance may remain in the dataset, and even identical images within a class may exist, leading to the generation of mislabeled pseudo labels for the FIQA regression network. Ultimately, this prevents the consistent learning of the regression network and hinders the model from reaching an optimal solution. Another limitation of conventional methods is that the training dataset for FR has a low proportion of low-quality image samples, making it difficult for regression networks to learn features of low-quality images.

To overcome these limitations, this study proposes two novel approaches to FIQA training that leverage sample relative classifiability. First, we propose to identify classes with low intra-class variance while training and assign them lower weights for the training loss. To identify classes with low intra-class variance, we utilize the exponential moving average (EMA) of the distance between the embedding and the prototype as an approximation of intra-class variance. The proposed method can effectively measure intra-class variance while requiring negligible computational resources and has the advantage of not requiring a pre-trained FR model. Second, we propose a novel and effective method to boost FIQA regression networks through on-the-fly data augmentation. The proposed method, IG-FIQA, leverages on-the-fly image rescaling, random erasing, and color jittering on training images, allowing the FIQA model to learn factors that may interfere with FR. This type of augmentation method poses the risk of impairing the performance of the FR model \cite{kim2022adaface}, potentially resulting in the generation of inaccurate pseudo-labels. Therefore, we have designed a method that safely incorporates data augmentation exclusively for FIQA regression network training. Our contributions can be summarized as follows:\\

\begin{itemize}
\item[$\bullet$] This paper introduces a novel approach to weight loss by incorporating Intra-class variance Guidance. This prevents the regression network from learning incorrect information through inappropriate pseudo-labels.
\item[$\bullet$] We propose a novel and effective method to boost FIQA regression networks via on-the-fly data augmentation to consider a variety of real face images.
\item[$\bullet$] IG-FIQA enables robust FIQA training and achieves state-of-the-art results on various benchmarks.
\end{itemize}

\section{Related work}

Existing FIQA methods can be classified into two types. One is to use embedding's properties, and the other is to predict face image quality using a regression network.

\subsection{Embedding's property based methods}

The embedding's properties based methods estimate the FIQ score by leveraging characteristics within the feature space or properties inherent to the facial recognition (FR) model. Probabilistic Face Embeddings (PFEs) \cite{shi2019probabilistic} proposed the method to represent a embedding as a Gaussian distribution in the latent space, where the mean of the distribution estimates the most likely feature values while the variance shows the uncertainty in the feature values. SER-FIQ \cite{terhorst2020ser} estimated the face image quality by calculating the distance between multiple embeddings on a query image, which were produced by different random subnetworks of the backbone. \cite{meng2021magface} and \cite{kim2022adaface} suggested a method to utilize the magnitude of embedding as an FIQ score, which is extracted from FR models trained with softmax-variant loss. \cite{deng2023harnessing} found that the feature distance between unrecognizable identity clusters and queries was correlated with the quality of face images, and used this distance as a FIQ score.

\subsection{Regression based methods}
Regression based FIQA approaches aim to train the regression network directly for predicting FIQ scores, unlike embedding property-based methods that do not require additional training. Given the absence of ground-truth data for face image quality, most methods within this approach aim to generate accurate pseudo-labels for image quality, facilitating the reliable training of regression networks. One easily devised method to obtain pseudo-labels is manual assignment by humans \cite{best2018learning}. FaceQnet \cite{hernandez2019faceqnet} proposed using the euclidean distance between the best quality image in the class and the target image as a pseudo-label. PCNet \cite{xie2020inducing} learned a face recognizer using only half of the dataset, then used half of the remaining dataset to construct a mated pair, and used the cosine similarity between pairs as a pseudo-label. SDD-FIQA \cite{ou2021sdd} proposed to use the distance between the intra-class similarity distribution and the inter-class similarity distribution as pseudo-labels. CR-FIQA \cite{boutros2023cr} proposed a method that utilizes the classifiability of embeddings as a pseudo-label. The network trained with pseudo-labels generated using classifiability has demonstrated its excellence by achieving state-of-the-art performance in various benchmarks.



\section{Methodology}

In this section, we explain the concepts and limitations of conventional SOTA method briefly and provide details of the proposed method to overcome these limitations by selectively assigning lower weight to samples belonging to classes with low intra-class variance and utilizing data augmentation.

\subsection{Revisiting CR-FIQA}
CR-FIQA \cite{boutros2023cr} is a FIQA method that utilizes relative classifiability as a pseudo-label for the FIQ score. In order to mathematically define classifiability, CR-FIQA introduces two novel concepts: Class Center Similarity (CCS) and Negative Nearest Class Center Similarity (NNCCS),
\begin{equation}
\begin{aligned}
    \mathrm{CCS}_{x_{i}} &= cos(\theta_{y_{i}}), \space\\
    \mathrm{NNCCS}_{x_{i}} &= \max_{j\in \left \{ 1,...,C \right \},j \neq y_{i}} cos(\theta_{j}),
\end{aligned}    
\end{equation}

where $C$ represents the total number of classes in the training dataset, while $y_{i}$ denotes the ground truth label corresponding to sample $x_{i}$. $\theta_{y_{i}}$ is the angle between the embedding $f(x_{i})$ extracted from the backbone network and the prototype $W_{y_{i}}$. $\mathrm{CCS}$ measures how similar an embedding is to the prototype of its corresponding class using cosine similarity, while $\mathrm{NNCCS}$ measure the similarity between an embedding and the prototype of the nearest negative class. Utilizing these two concepts, pseudo-labels to train a regression network is defined as follows:
\begin{equation}
\label{cr_equation}
    \mathrm{CR}_{x_{i}} = \frac{\mathrm{CCS}_{x_{i}}}{\mathrm{NNCCS}_{x_{i}} + (1+\epsilon)},
\end{equation}
where $\epsilon$ is set to $1e-9$ to prevent division by zero. According to its definition, a $\mathrm{CR}_{x_{i}}$ value increases as an embedding approaches the positive class prototype and diverges from prototypes of nearest negative classes. Consequently, a higher $\mathrm{CR}_{x_{i}}$ value implies that the sample $x_{i}$ is more easily classifiable. 

CR-FIQA employs $\mathrm{CR}$ as pseudo-labels to train a regression network, $R \in D\times1$, which consists of a single linear layer taking the embedding feature with dimension $D$ from the FR backbone as input. Smooth L1 loss was used to avoid gradient explosion. For each sample, its loss is defined as,
%
%
\begin{equation}
l_{\mathrm{CR}}\left ( x_{i} \right )=\left\{\begin{matrix}
 0.5/\beta \times \left (d_{\mathrm{CR}}\left ( x_{i} \right )\right )^2 & {\scriptstyle \text{if } \left | d_{\mathrm{CR}}\left ( x_{i} \right ) \right | < \beta} \\
\left | d_{\mathrm{CR}}\left ( x_{i} \right ) \right | - 0.5 \times \beta & {\scriptstyle\text{otherwise}} \\
\end{matrix}\right.,\\
\end{equation}
where $d_{\mathrm{CR}}\left ( x_{i} \right )=\mathrm{CR}_{x_{i}} - R\left(f\left(x_{i}\right)\right)$. The total loss for training CR-FIQA is a combination of $L_{\mathrm{Arc}}$, which trains the backbone network as in~\cite{deng2019arcface}, and $L_{\mathrm{CR}}$, which trains the regression network:
\begin{equation}
L_{\mathrm{CR}}=\sum_{i}l_{\mathrm{CR}}\left(x_i  \right),
\end{equation}
\begin{equation}
L_{\mathrm{CR-FIQA}} = L_{\mathrm{Arc}} + \lambda \times L_{\mathrm{CR}}.
\end{equation}
$\lambda$ is weight parameter to balance angular margin loss $L_{\mathrm{Arc}}$ and regression loss $L_{\mathrm{CR}}$, and set to 10.0 in CR-FIQA.

\begin{figure*}[t!]
	\center
	\includegraphics[width=0.95\linewidth]{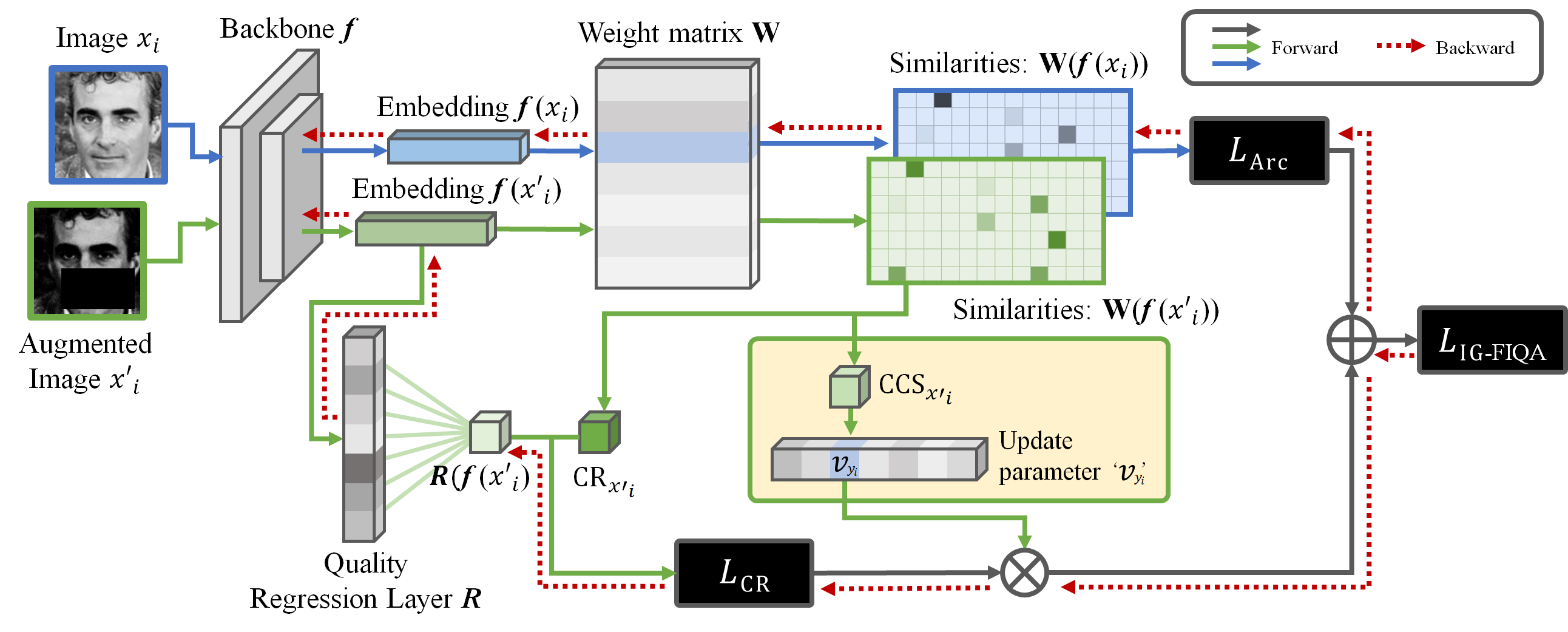}
	\caption{
		An overview of IG-FIQA training process. IG-FIQA utilizes an approximation of the intra-class variance to handle the adverse effects of samples with low intra-class variance. Note that the weight parameter $v_{i}$ does not require gradient updates during training. Original image forward pass: \textbf{\textcolor{frenchblue}{Blue}}, augmented image forward pass: \textbf{\textcolor{kellygreen}{Green}}.
	}
	\label{overview}
\end{figure*}

\subsection{IG-FIQA}
%
\textbf{Mitigating the impact of low intra-class variance.}
To handle the impact of images with low intra-class variance on the pseudo-labels for FIQ, we investigate approaches to identify classes exhibiting low intra-class variance during training. This process involves computing intra-class variance \cite{pilarczyk2019intra}, defined as follows: 
%
%
\begin{equation}
\label{variance_equation}
\begin{aligned}
    var_{y_{i}} &= \frac{1}{N}\sum_{i}^{N} \left \| f(x_{i})- \mu_{f(x)} \right \|^{2},\\
    \mu_{f(x)} &= \frac{1}{N}\sum_{i}^{N} f(x_{i}),
\end{aligned}
\end{equation}
%
%
%
where $N$ represents the number of samples in the class $y_{i}$. Computing $var_{y_{i}}$ at every iteration is a straightforward approach to identify classes with low intra-class variance. However, calculating the class variance every iteration with this primitive method is highly inefficient and practically infeasible. For efficient computation, we propose a method to approximate $var_{y_{i}}$ and $\mu_{f(x)}$ fairly accurately. Firstly, we leverage the observation that as training progresses, prototype $W_{y_{i}}$ converges to the class centroid $\mu_{f(x)}$. At this point, $\left \| f(x_{i})- \mu_{f(x)} \right \|^{2}$ term in ~\cref{variance_equation} could be approximated by $1-\mathrm{CCS}_{x_{i}}$. With this approximation, $var_{y_{i}}$ can be represented as follows:

\begin{equation} \label{approx1}
    var_{y_{i}} \approx \frac{1}{N}\sum_{i}^{N} 1-\mathrm{CCS}_{x_{i}},
\end{equation}

With the suggested approximation ~\cref{approx1}, we no longer need to calculate $\mu_{f(x)}$ in order to compute the class variance. To further simplify the calculation of the average of $1-\mathrm{CCS}_{x_{i}}$, we utilized the exponential moving average (EMA). By this, the intra-class variance $v_{y_{i}}^{t}$ for class $y_{i}$ at the $t$ th iteration can be represented by the following:
%
\begin{equation}
var_{y_{i}} \approx v_{y_{i}}^{t} = \alpha \times v_{y_{i}}^{t-1} + (1-\alpha) \times (1-\mathrm{CCS}_{x_{i}}),
\end{equation}
%
where $\alpha$ is a momentum hyperparameter. If $\alpha$ is small, $var_{y_{i}}$ can be greatly affected by $\mathrm{CCS}_{x_{i}}$, and conversely, if $\alpha$ is small, $var_{y_{i}}$ will be affected less. The $\mathrm{CCS}$ values undergo significant fluctuations in the early stages of learning since the model parameters has not fully converged, while the variation becomes minor in the later stages of training. For this reason, we gradually increased the $\alpha$ from 0.9 to 1.0 until the last epoch $e_{end}$ of training. Detailed experiment and analysis for hyperparameter $\alpha$ is descibed in the ablation study \ref{ablation}. Since $\mathrm{CCS}_{x_{i}}$ should be computed at every iteration to generate $\mathrm{CR}_{x_{i}}$, the proposed method has negligible computational burden on computing $v_{y_{i}}$.

Afterwards, $v_{y_{i}}$'s are adjusted to the range of [0, 1] through z-score normalization:
\begin{equation}
\left \| \hat{v_{y_{i}}} \right \| = 1+ \left \lfloor \frac{v_{y_{i}} - \mu_{v}}{\sigma_{v}} \right \rceil_{-1}^{0},
\end{equation}
where $\mu_{v}$ and $\sigma_{v}$ are the mean and standard deviation of $v_{y_{i}}$ computed across all classes, respectively. As a result, $\left \| \hat{v_{y_{i}}} \right \|$ is intended to be $1$ when the class $y_{i}$ comprises diverse images, and tends towards $0$ in the case of homogeneous and similar images. This value of $\left \| \hat{v_{y_{i}}} \right \|$ serves as the weight parameter for $L_{CR}$ during regression network training. Since approximately 16$\%$ of the unit gaussian distribution has values less than -1, IG-FIQA trains using only classes with intra-class variance in the top $84\%$.
\begin{equation}
    L_{\mathrm{IG}} = \sum_{i} \left \| \hat{v_{y_{i}}} \right \| \times l_{\mathrm{CR}}(x_{i}),
\end{equation}

$v_{y_{i}}$'s are initialized to 1.0 for all classes at the beginning of training so that all data samples could equally contribute to training. The overall loss of the proposed method can be represented as follows:
\begin{equation}
    L_{\mathrm{IG-FIQA}} = L_{\mathrm{Arc}} + \lambda \times L_{\mathrm{IG}}.
\end{equation}

We set $\lambda$ to 10.0, following the original CR-FIQA. Further experiments detailed in \ref{ablation} demonstrate that the proposed method can calculate the class variance fairly accurately, taking only \textbf{0.7 seconds} per iteration, whereas the naive approach took \textbf{58.7 seconds} on RTX3090 with the CASIA-WebFace dataset and a mini-batch size of 1024.\\\\
\textbf{Boosting FIQA through data augmentation.} The proposed IG-FIQA applies rescaling, random erasing, and color jittering as data augmentations that could degrade the face image quality. This augmentation improves the model's adaptability to a variety of low-quality facial images that could exist in the unconstrained real world, such as images acquired from CCTV. However, using heavily degraded face images for training runs the risk of overfitting the FR backbone to extract features from non-face information, which may ultimately hinder the FR backbone training \cite{kim2022adaface}. For this reason, we do not utilize the augmented data for $L_{\mathrm{Arc}}$ calculation but use it for calculating the loss only for the regression network. This is achieved through a simple mini-batch separation, allocating one part for the regression network and the other for backbone network training. Our pipeline is specifically designed to ensure that augmentations degrading image quality are excluded from the batch forwarded to the FR backbone network, thereby preserving the integrity of the training process. The complete forward pass and backward pass of the proposed method can be seen in the overview Fig. \ref{overview}. As shown in the overview, quality-degraded facial images are only used for calculating the regression loss $L_{\mathrm{IG}}$ and generating the pseudo-label $\mathrm{CR}{x_{i}}$.\\

\begin{figure}[t]
\center
\begin{subfigure}{0.61\linewidth}
\includegraphics[width=\linewidth]{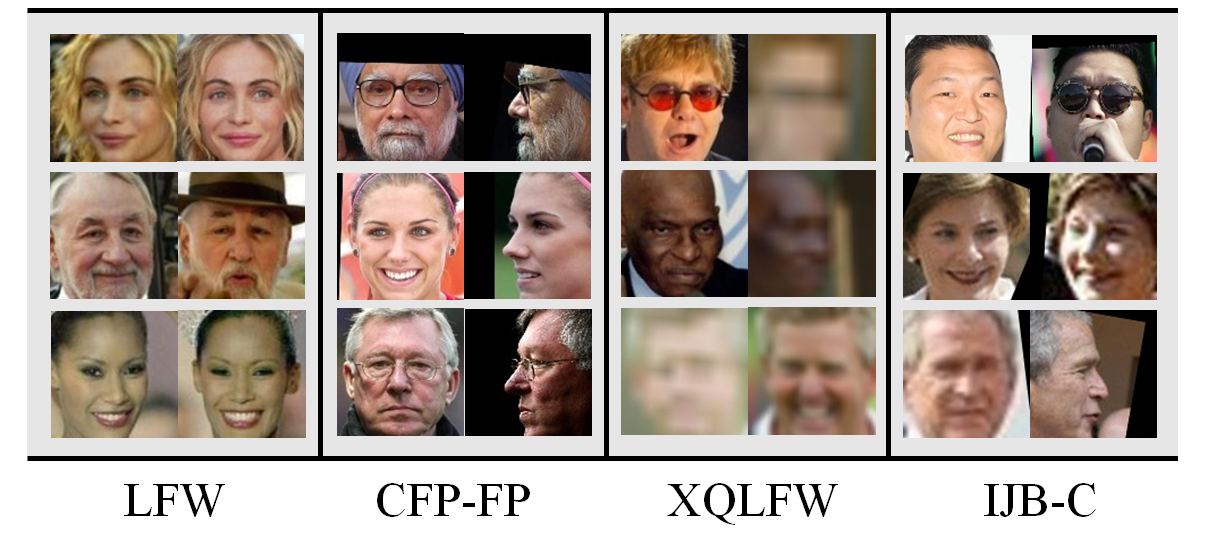}
\caption{Examples of facial image in the benchmark dataset.}
\label{four_bench_example}
\end{subfigure}
\begin{subfigure}{0.38\linewidth}
\includegraphics[width=\linewidth]{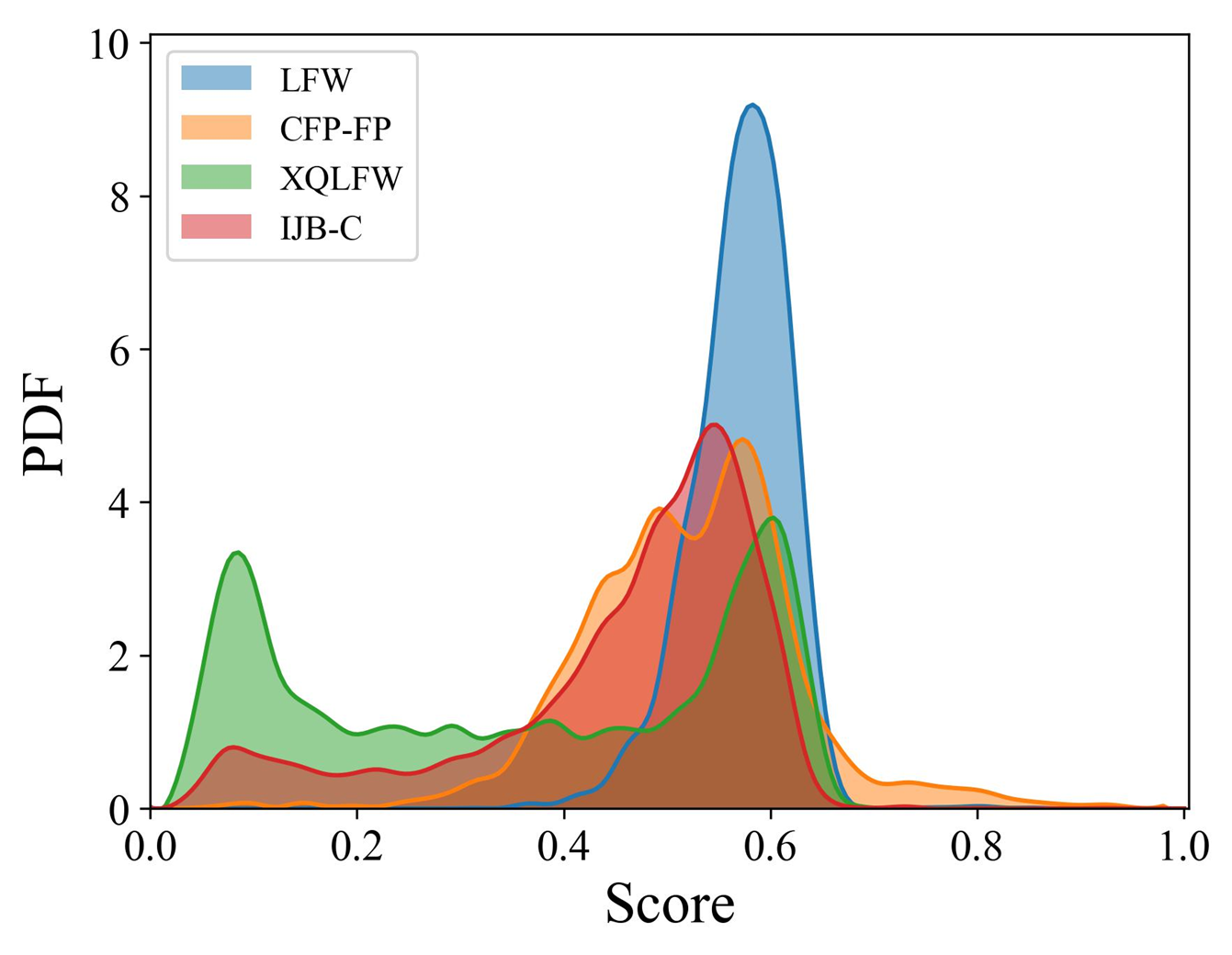}
\caption{Quality score distribution of benchmark datasets.}
\label{four_bench_distribution}
\end{subfigure}
\caption{In Fig. \ref{four_bench_example}, pictures in the same row belong to the same ID. The quality scores for Fig. \ref{four_bench_distribution} were obtained using IG-FIQA and have been normalized to [0, 1].}
\vspace*{-3mm}

\end{figure}

\section{Experiments and Results}
\subsection{Implementation Details}
\indent\textbf{Datasets.} We utilized the CASIA-WebFace \cite{yi2014learning} and MS1M-V2 \cite{deng2019arcface} datasets for training. For evaluation, we employed the LFW \cite{huang2008labeled}, CFP-FP \cite{sengupta2016frontal}, CPLFW \cite{zheng2018cross}, XQLFW \cite{knoche2021cross}, IJB-B \cite{whitelam2017iarpa}, and IJB-C \cite{maze2018iarpa} datasets. All images used in training and evaluation were cropped and aligned to a size of $112 \times 112$ pixels as specified in ~\cite{liu2017sphereface,wang2018cosface}. While LFW, CFP-FP, and CPLFW are extensively used benchmarks, the performance of FR models has reached a saturation point due to the predominance of high-quality images in these datasets \cite{kim2022adaface}. On the other hand, XQLFW, IJB-B, and IJB-C are benchmarks consisting of a mixture of high-quality and low-quality images. This indicates that they are suitable datasets for evaluating FIQA performance. Overall, these evaluation datasets cover various challenges for FR, including variations in pose, illumination, and resolution. For a better understanding of the image qualities within the benchmarks, we plot the FIQA score distribution using IG-FIQA in Fig. \ref{four_bench_distribution}. Examples of facial image quality for LFW, CFP-FP, XQLFW, and IJB-C can be found in Fig. \ref{four_bench_example}.
\\\\\textbf{Experiment Settings.} Similar to CR-FIQA, the performance evaluation of the proposed IG-FIQA was conducted under two distinct protocols: a small protocol (IG-FIQA(S)) and a large protocol (IG-FIQA(L)). In IG-FIQA(S), we utilized ResNet-50 as the backbone and the CASIA-WebFace as training dataset. We set the initial learning rate to 1e-1, divide the learning rate by 10 at 20 and 28 epochs, and end training after 36 epochs. For IG-FIQA(L), ResNet-100 used as the backbone and MS1M-V2 used as the training dataset. The initial learning rate was set to 1e-1 and divided by 10 in 10 and 16 epochs, and training was ended after 20 epochs. For both protocols, the SGD optimizer with a momentum of 0.9 and weight decay of 5e-4 was employed. Regarding the ArcFace loss, the scale parameter (s) and the margin (m) remained at 64.0 and 0.5, respectively, following the specifications from the original paper \cite{deng2019arcface}. We set the mini-batch size to 1024, with 512 images for regression network training and the remaining 512 images for FR backbone training.
\\\\\textbf{Evaluation metrics.} The Error versus Rejection Curve (ERC), which is the most common method for measuring FIQA performance~\cite{grother2007performance,grother2014face}, was used to compare the performance with recent SOTA FIQA methods. It measures verification performance through False None Match Rate (FNMR) based on the rejection rate of the quality score at a fixed False Match Rate (FMR). Additionally, we reported the Area Under Curve (AUC) of ERC in Tab. \ref{bench_table} to evaluate verification performance across all rejection rate intervals of the ERC. A smaller AUC value indicates better performance of the FIQA model. All the experimental results presented in this paper were obtained under cross-model settings; the FIQA models were solely employed to assess the quality of face images, and the embedding features were extracted using independent pre-trained FR models.
\\\\\textbf{Augmentations.} In this paper, we simply adopt rescaling, random erasing, and color jittering as augmentations to train the FIQA regression network. These methods are commonly used to train classification networks and intentionally degrade image quality \cite{he2019bag}. Specifically, rescaling involved shrinking the image and then restoring it to the original size, resulting in blurring of the face image. For random erasing, we randomly selected a rectangular area from the sample and set its pixel values to 0. Color jittering randomly modified the brightness, contrast, and saturation of the image. Additionally, random horizontal flip was applied to both mini-batches forwarded to the FR backbone and regression network training, as it does not degrade the image quality.
\\\\\textbf{Face Recognition Models.} In the experiment, we used six commonly used models for FR: CosFace \cite{wang2018cosface}, ArcFace \cite{deng2019arcface}, CurricularFace \cite{huang2020curricularface}, MagFace \cite{meng2021magface}, ElasticFace \cite{boutros2022elasticface} and AdaFace \cite{kim2022adaface}, all trained with ResNet-100 as the backbone using MS1M-V2 dataset. In our experiments, all FR models utilized pre-trained weights available in the official repository, except for CosFace \cite{wang2018cosface} and ArcFace \cite{deng2019arcface}, which were re-implemented due to the lack of pre-trained models under the same conditions.
%
%
%
\begin{figure}[t]
\center
\begin{subfigure}{0.35\linewidth}
\includegraphics[width=\linewidth]{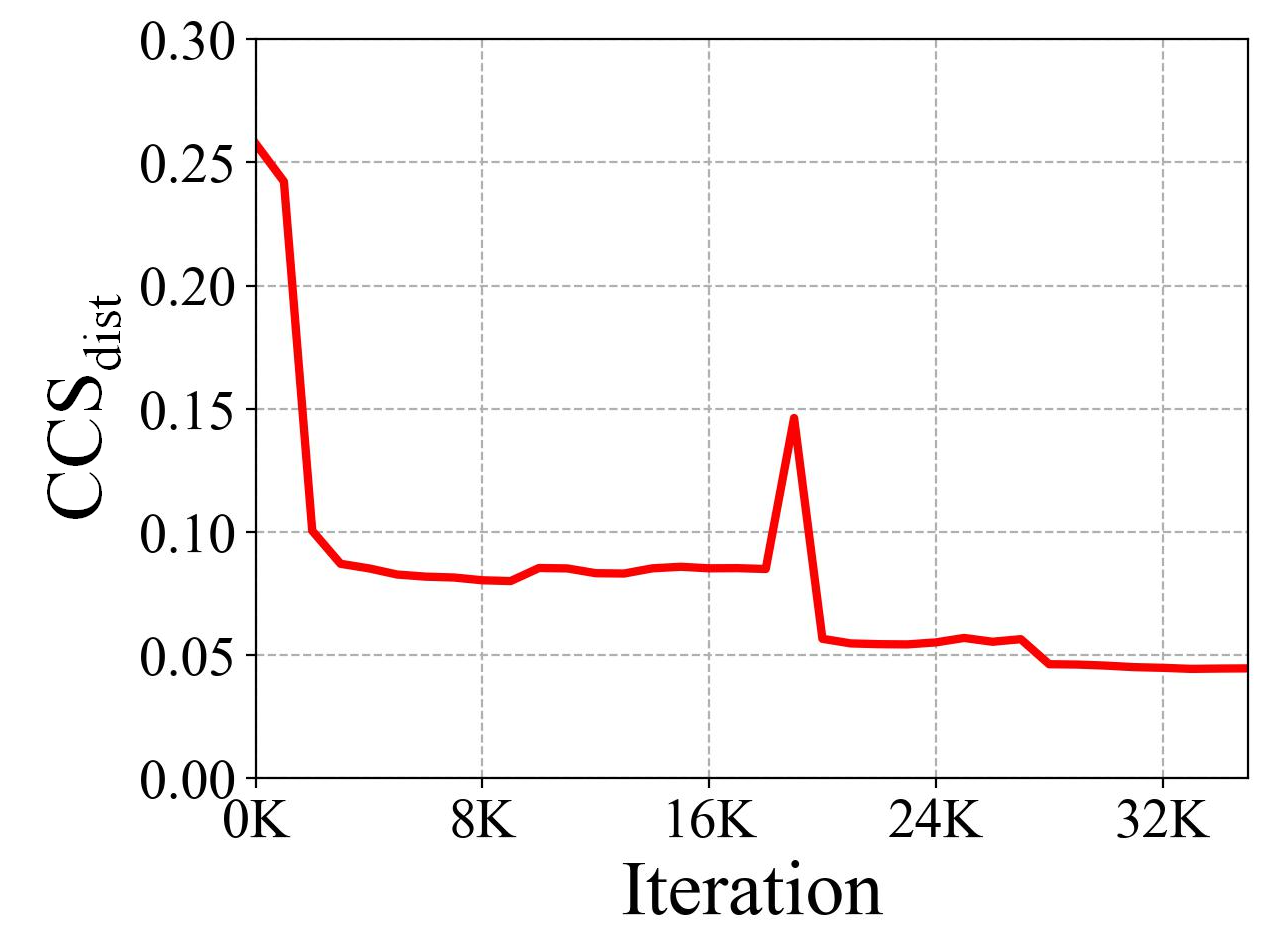}
\caption{Changes of $\mathrm{CCS}_{\mathrm{dist}}$ during training.}
\label{correlation1}
\end{subfigure}
\quad\quad
\begin{subfigure}{0.35\linewidth}
\includegraphics[width=\linewidth]{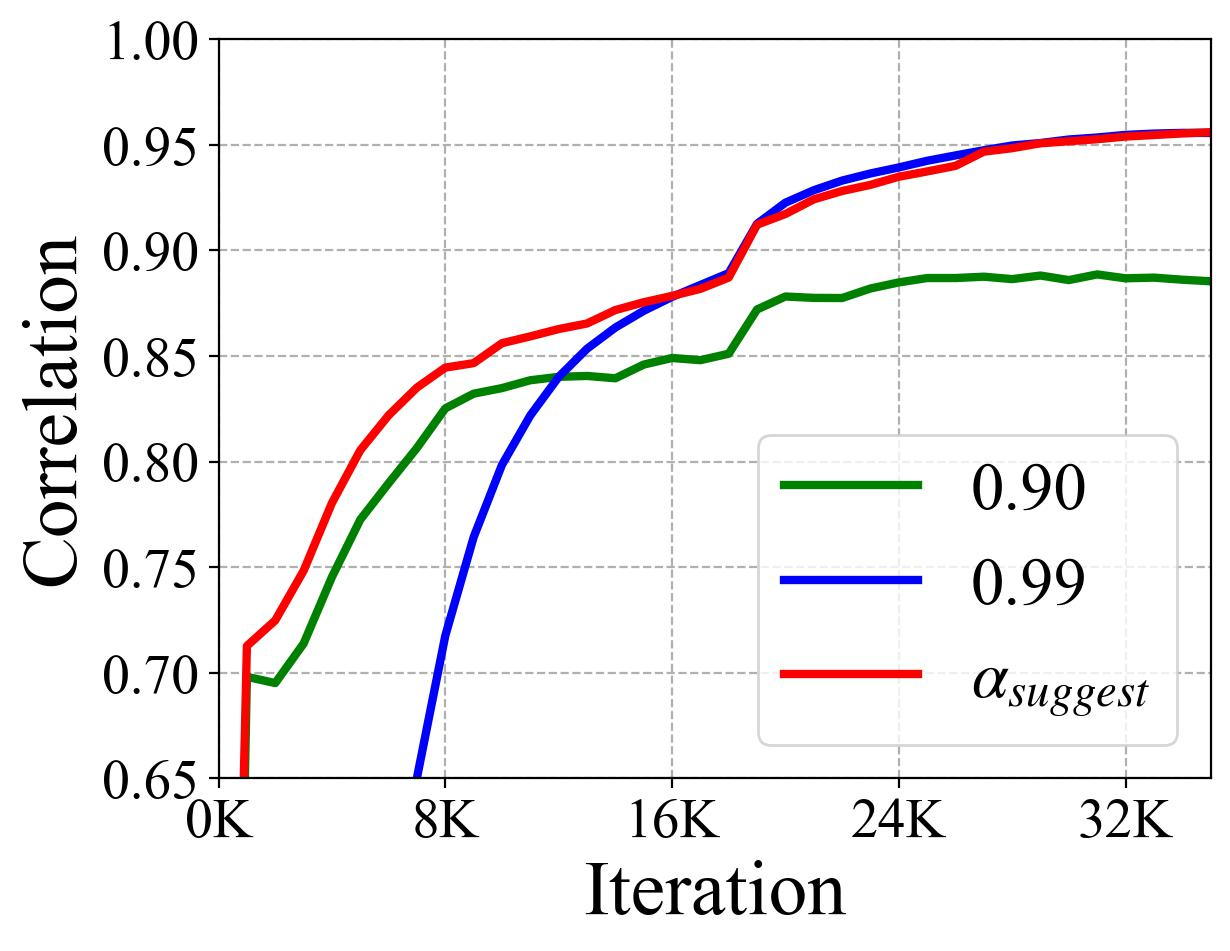}
\caption{Changes of pearson correlation $\rho_{var, v}$ during training.}
\label{Correlation2}
\end{subfigure}        
\caption{Ablation study on momentum paramter $\alpha$.}
\end{figure}
\subsection{Ablation and analysis} \label{ablation}
\textbf{Effect of momentum paramter $\alpha$.} The hyperparameter $\alpha$ is an important factor that determines how much the $\mathrm{CCS}_{x_{i}}$ in the current iteration influences the weight parameter $\left \| \hat{v} \right \|$. More specifically, the influence of the $\mathrm{CCS}_{x_{i}}$ in the current iteration on the weight parameter is inversely proportional to the $\alpha$. To determine an appropriate setting for the $\alpha$, we tracked the average change of $\mathrm{CCS}_{x_{i}}$ ($\mathrm{CCS}_{\mathrm{dist}}$) throughout each epoch of the training process, as follows:
\begin{equation}
    \mathrm{CCS}_{\mathrm{dist}} = \frac{1}{k}\sum_{i=1}^{k}\left | \mathrm{CCS}_{x_{i},e_{t}}-\mathrm{CCS}_{x_{i},e_{t-1}} \right |,
\end{equation}
where $\mathrm{CCS}_{x_{i},e_{t}}$ refers to the $\mathrm{CCS}$ obtained using the $x_{i}$ embedding after the $t$ th epoch $e_{t}$. As shown in Fig. \ref{correlation1}, $\mathrm{CCS}_{\mathrm{dist}}$ is large in the early stages of training, but gradually decreases as the model converges. In order to quickly reflect changing CCS values in the weight parameters, it is more advantageous to use a low momentum parameter. Conversely, in the later stages of learning, it is reasonable to design the weight parameter to be less affected by the instance $\mathrm{CCS}_{x_{i}}$ by using a high $\alpha$. For this reason, we use low momentum parameters at the beginning of training and gradually increase the momentum parameters to 1.0 until the end of training.

To verify the efficacy of the proposed variable $v_{y_{i}}$, we calculated the correlation between $v_{y_{i}}$ and the intra-class variance $var_{y_i}$ during training. Intra-class variance $var_{y_{i}}$ is computed with pretrained ResNet-50 ArcFace model. Fig. \ref{Correlation2} shows the changes in the Pearson correlation coefficient $\rho_{var, v}$ between $var_{y_{i}}$ and $v_{y_{i}}$. As depicted in the figure, the correlation between the two variables is maximized when the momentum parameter $\alpha$ gradually increases from 0.9 to 1.0 during training. We also observed that the Pearson correlation between $var_{y_{i}}$ and the proposed $v_{y_i}$ reaches a high value ($>0.71$) from the end of the second training epoch. This indicates that the proposed method measures intra-class variance fairly accurately from the early stage of training and reflects it in regression network training.
\begin{figure*}[t]
	\center
	\includegraphics[width=0.8\linewidth]{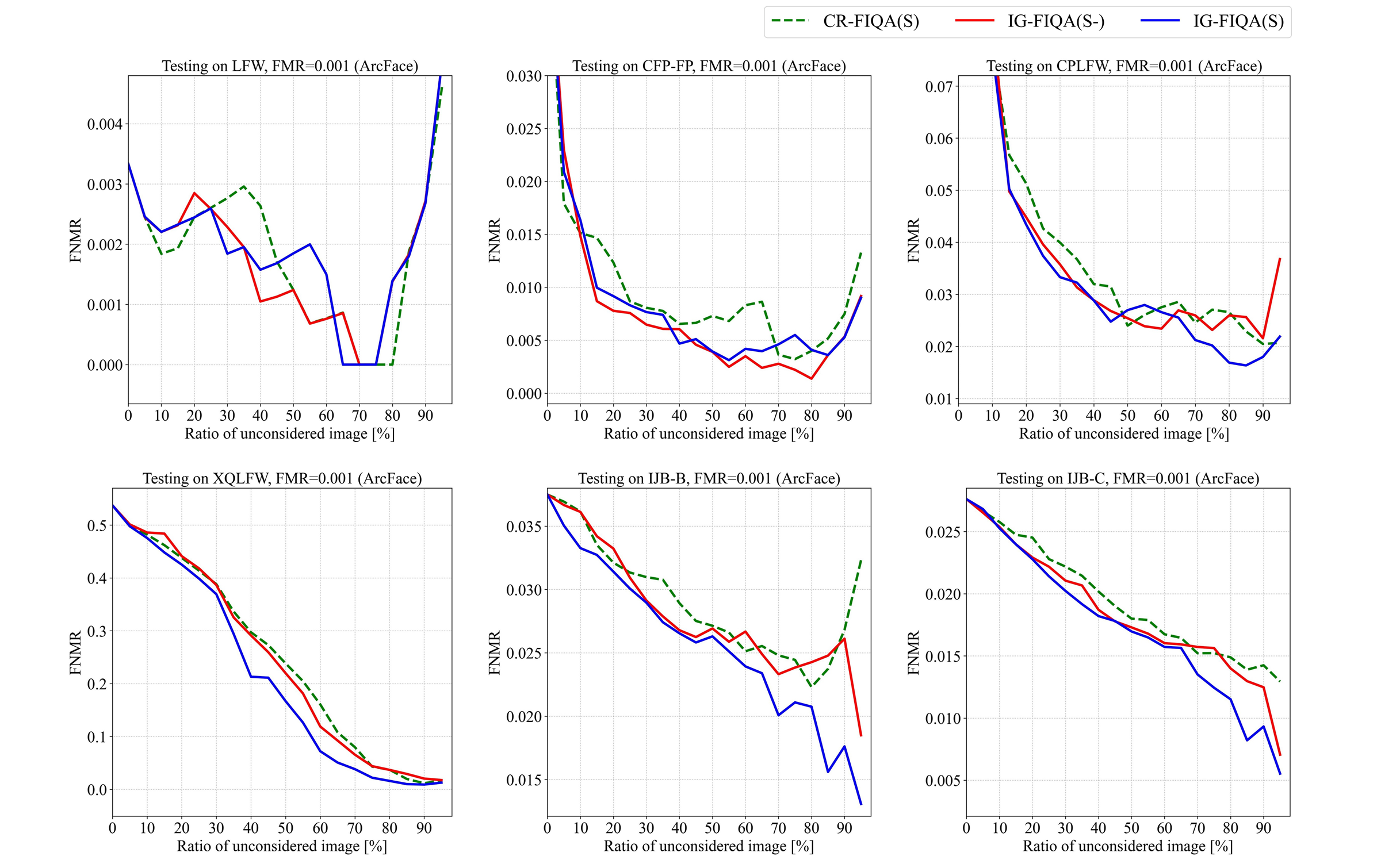}
	\caption{
        ERC plots comparing conventional SOTA method, our method without augmentation (IG-FIQA(S-)), and our method with augmentation (IG-FIQA(S)).
	}
	\label{aug_ablation}
\end{figure*}
\\\\\textbf{Effect of }$L_{\mathrm{IG}}$\textbf{ loss.}
To validate the effectiveness of the proposed weight parameter, we plotted the ERC of IG-FIQA(S) without augmentation, IG-FIQA(S) with data augmentation, and CR-FIQA(S) in Fig. \ref{aug_ablation}. As shown in the figure, we can see that even without data augmentation, IG-FIQA(S) outperforms CR-FIQA(S) on both high-quality and mixed-quality datasets. This result shows that ignoring classes with low intra-class variance during training is effective for model generalization. In Fig. \ref{weight_average}, we plot the distribution of the weight parameter $\left \| \hat{v} \right \|$ assigned to each class in MS1M-V2 dataset after training. Below the distribution, we present the average face image of the class corresponding to the distribution. As can be seen in the Fig. \ref{weight_average}, 16\% of classes with $\left \| \hat{v} \right \|$ equal to 0 are ignored for the training of the regression network. The average facial image derived from these classes looks like a single image, due to low intra-class variance.
\begin{table}[t]
\begin{subtable}[h]{0.50\textwidth}
\centering
\resizebox{\linewidth}{!}{%
\label{weight_real}
\begin{tabular}{c|c|c|c|c|c|c|c} \hline
FR & Data-aug & LFW & CFP-FP & CPLFW & XQLFW & IJB-B & IJB-C \\ \hline\hline
\multirow{4}{*}{ArcFace} & - & \textcolor{red}{\textbf{0.0016}} & \textbf{\textcolor{red}{0.0070}} & 0.0396 & 0.2339 & 0.0268 & 0.0177 \\
 & 20\% & 0.0017 & 0.0083 & 0.0396 & 0.2206 & 0.0246 & 0.0168 \\
 & \textbf{30\%} & 0.0017 & 0.0077 & \textcolor{red}{\textbf{0.0374}} & \textbf{\textcolor{red}{0.2059}} & \textcolor{red}{\textbf{0.0245}} & 0.0166 \\
 & 40\% & 0.0017 & 0.0081 & 0.0387 & 0.2147 & 0.0246 & \textcolor{red}{\textbf{0.0164}} \\ \hline
\multirow{4}{*}{AdaFace} & - & 0.0019 & \textcolor{red}{\textbf{0.0091}} & 0.0341 & 0.1708 & 0.0225 & 0.0143 \\
 & 20\textbf{\textbf{\%}} & \textcolor{red}{\textbf{0.0017}} & 0.0099 & 0.0341 & \multicolumn{1}{l|}{0.1522} & 0.0205 & 0.0133 \\
 & \textbf{30\%} & 0.0018 & 0.0098 & \textbf{\textcolor{red}{0.0323}} & \textcolor{red}{\textbf{0.1432}} & \textcolor{red}{\textbf{0.0205}} & \textcolor{red}{\textbf{0.0132}} \\
 & 40\textbf{\textbf{\%}} & 0.0018 & 0.0094 & 0.0336 & 0.1434 & 0.0208 & 0.0132 \\ \hline
\end{tabular}
}
\caption{The AUCs of ERCs in FMR=1e-3, according to augmentation ratio. \color{red}\textbf{Red} \color{black}is the best.}
\label{augmentation_ratio}
\end{subtable}
\hfill
\begin{subtable}[h]{0.50\textwidth}
\centering
\resizebox{\linewidth}{!}{%
\label{weight_vir}
\begin{tabular}{l|c|c|c|c|c|c|c} \hline
Methods & Data-aug & LFW & CFP-FP & CPLFW & XQLFW & IJB-B & IJB-C \\ \hline\hline
CR-FIQA(S) & - & 99.35 & 96.59 & 85.30 & 66.23 & 77.72 & 82.10 \\
CR-FIQA(S) & \checkmark & 98.63 & 82.10 & 75.70 & 67.70 & 64.75 & 63.21 \\ \hline
IG-FIQA(S) & - & 99.35 & 96.04 & 86.00 & 69.45 & 83.07 & 85.19 \\
IG-FIQA(S) & \checkmark & 99.38 & 96.37 & 86.90 & 68.22 & 82.27 & 85.96 \\ \hline
\end{tabular}
}
\caption{Performance degradation of FR backbones depending on data augmentation. Verification accuracy for IJB-B and IJB-C are reported on TAR@FAR=1e-3.}
\label{augmentation_FR}
\end{subtable}
\caption{Ablation study for augmentations.}
\label{tab_weight}
\vspace*{-6mm}

\end{table}
%
%
\\\\\textbf{Ablation study on augmentation.}
To find the optimal data augmentation ratio, we trained the IG-FIQA(S) using various data augmentation ratios. Tab. \ref{augmentation_ratio} shows the AUC of ERC evaluated on various benchmarks at FMR=1e-3. As seen in the table, we observed that the model trained using an augmentation ratio of $30\%$ achieved the best performance in most cases. Based on this experiment, we applied rescaling, random erasing, and color jittering at a rate of $30\%$  probability each, resulting only 34.3\% $(0.7^{3})$ of images statistically not undergoing any augmentation during training. As can be seen in Fig. \ref{aug_ablation}, IG-FIQA(S) with data augmentation achieves a further performance improvement than its non-augmented counterpart, especially in mixed-quality benchmarks.

To demonstrate that the suggested pipeline effectively protects the FR backbone from the risk of degradation due to augmentation, we measured the FR verification accuracy of the trained FIQA backbone in Tab. \ref{augmentation_FR} using FR benchmarks. As observed in the table, the proposed separated pipeline method enabled stable backbone training regardless of augmentation, while the original CR-FIQA backbone suffered performance degradation in most benchmarks due to data augmentations. This indicates that the proposed pipeline, with batch separation, can effectively prevent performance degradation caused by data augmentations, thereby ensuring stable training of the FR backbone, which is necessary for accurate pseudo-label generation.
\begin{figure*}[!ht]
\center
\includegraphics[width=0.88\linewidth]{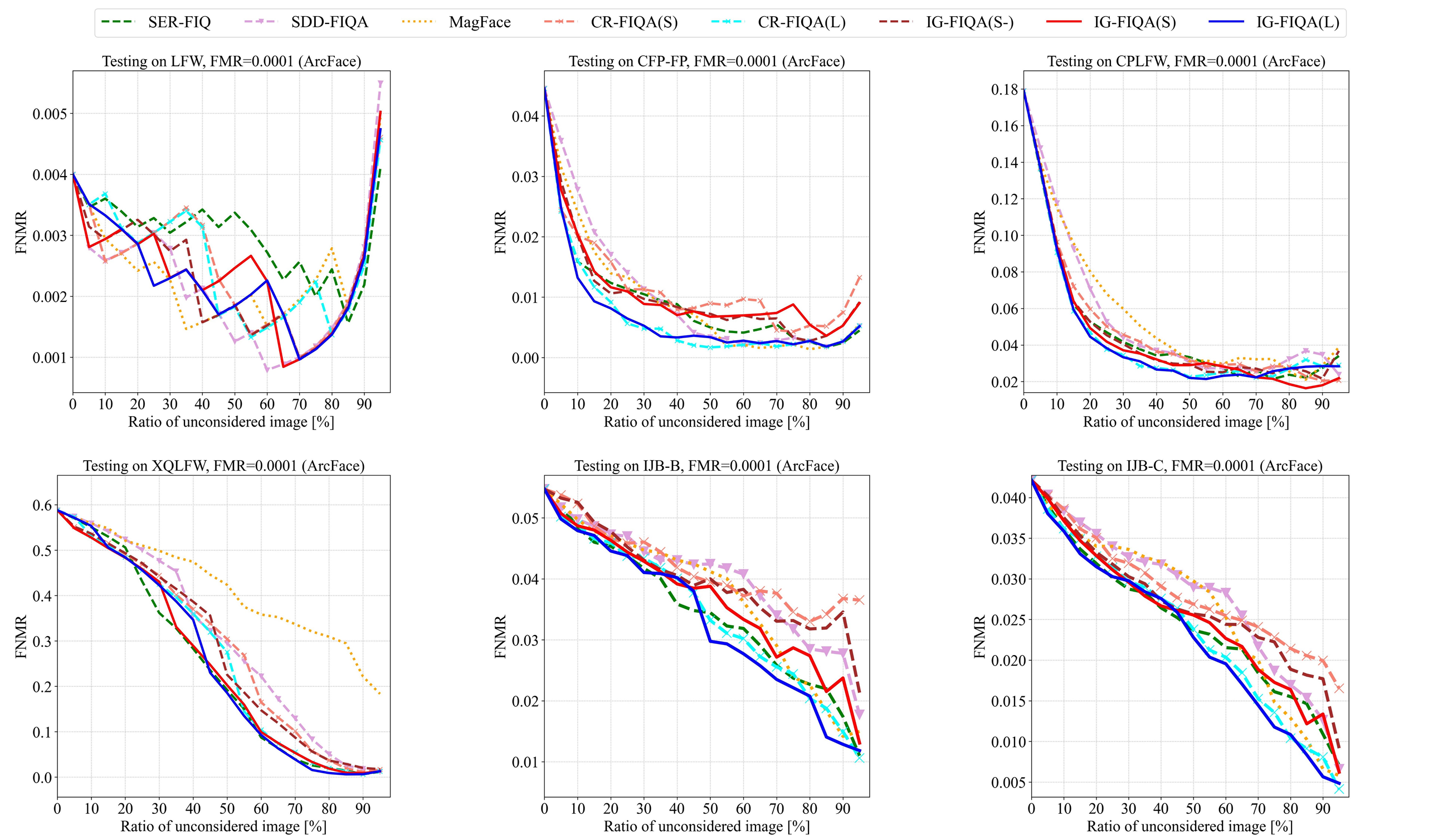}
\vspace*{-2mm}
\caption{
ERC plots on ArcFace.
}
\label{erc_plot1}
\vspace*{-13mm}
\end{figure*}
\begin{figure*}[!h]
\center
\includegraphics[width=0.88\linewidth]{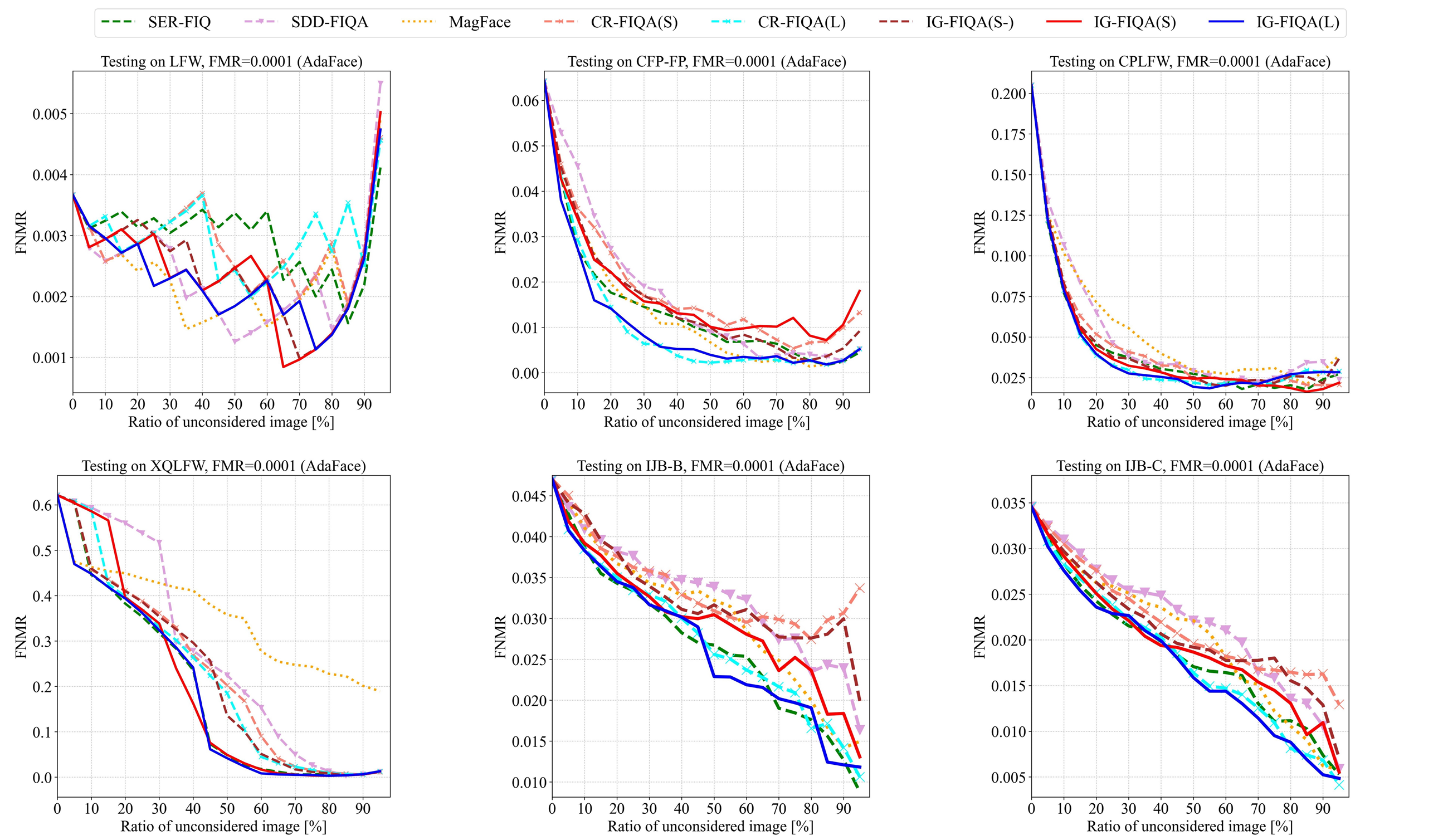}
\vspace*{-2mm}
\caption{
ERC plots on AdaFace.
}
\label{erc_plot2}
\vspace*{-10mm}
\end{figure*}
\begin{table*}[!ht]
\centering
\resizebox{1.0\textwidth}{!}{%
\begin{tabular}{c|c|cc|cc|cc|cc|cc|cc} \hline
\multirow{3}{*}{FR} & \multirow{3}{*}{Method} & \multicolumn{6}{c|}{High-quality} & \multicolumn{6}{c}{Mixed-quality} \\ \cline{3-14}
 &  & \multicolumn{2}{c|}{LFW} & \multicolumn{2}{c|}{CFP-FP} & \multicolumn{2}{c|}{CPLFW} & \multicolumn{2}{c|}{XQLFW} & \multicolumn{2}{c|}{IJB-B} & \multicolumn{2}{c}{IJB-C} \\ \cline{3-14}
 &  & 1e-3 & 1e-4 & 1e-3 & 1e-4 & 1e-3 & 1e-4 & 1e-3 & 1e-4 & 1e-3 & 1e-4 & 1e-3 & 1e-4 \\ \hline\hline
\multirow{8}{*}{CosFace†} & SER-FIQ & 0.0021 & 0.0025 & 0.0100 & 0.0183 & 0.0404 & 0.0462 & \textbf{\textcolor{blue}{0.2128}} & 0.2580 & 0.0246 & \textcolor{blue}{\textbf{0.0370}} & 0.0166 & 0.0272 \\
 & SDD-FIQA & \textbf{\textcolor{red}{0.0012}} & 0.0020 & 0.0096 & 0.0185 & 0.0464 & 0.0518 & 0.2607 & 0.2937 & 0.0297 & 0.0447 & 0.0196 & 0.0310 \\
 & MagFace & 0.0014 & \textcolor{blue}{\textbf{0.0019}} & 0.0096 & 0.0169 & 0.0496 & 0.0545 & 0.3997 & 0.4823 & 0.0272 & 0.0414 & 0.0181 & 0.0290 \\
 & CR-FIQA(S) & 0.0016 & 0.0023 & 0.0098 & 0.0240 & 0.0412 & 0.0472 & 0.2331 & 0.3156 & 0.0306 & 0.0460 & 0.0198 & 0.0318 \\
 & CR-FIQA(L) & 0.0017 & 0.0022 & \textcolor{blue}{\textbf{0.0075}} & \textcolor{blue}{\textbf{0.0130}} & \textcolor{red}{\textbf{0.0373}} & \textcolor{blue}{\textbf{0.0429}} & 0.2133 & \textcolor{blue}{\textbf{0.2466}} & \textcolor{blue}{\textbf{0.0245}} & 0.0371 & \textcolor{blue}{\textbf{0.0159}} & \textcolor{blue}{\textbf{0.0258}} \\
 & IG-FIQA(S-)(Our) & 0.0016 & 0.0022 & 0.0097 & 0.0213 & 0.0402 & 0.0461 & 0.2418 & 0.2962 & 0.0294 & 0.0443 & 0.0188 & 0.0303 \\
 & {\cellcolor[rgb]{0.937,0.937,0.937}}IG-FIQA(S)(Our) & {\cellcolor[rgb]{0.937,0.937,0.937}}0.0015 & {\cellcolor[rgb]{0.937,0.937,0.937}}0.0022 & {\cellcolor[rgb]{0.937,0.937,0.937}}0.0100 & {\cellcolor[rgb]{0.937,0.937,0.937}}0.0210 & {\cellcolor[rgb]{0.937,0.937,0.937}}0.0375 & {\cellcolor[rgb]{0.937,0.937,0.937}}0.0433 & {\cellcolor[rgb]{0.937,0.937,0.937}}0.2179 & {\cellcolor[rgb]{0.937,0.937,0.937}}0.2560 & {\cellcolor[rgb]{0.937,0.937,0.937}}0.0270 & {\cellcolor[rgb]{0.937,0.937,0.937}}0.0403 & {\cellcolor[rgb]{0.937,0.937,0.937}}0.0177 & {\cellcolor[rgb]{0.937,0.937,0.937}}0.0280 \\
 & {\cellcolor[rgb]{0.937,0.937,0.937}}IG-FIQA(L)(Our) & {\cellcolor[rgb]{0.937,0.937,0.937}}\textcolor{blue}{\textbf{0.0013}} & {\cellcolor[rgb]{0.937,0.937,0.937}}\textbf{\textcolor{red}{0.0018}} & {\cellcolor[rgb]{0.937,0.937,0.937}}\textbf{\textcolor{red}{0.0070}} & {\cellcolor[rgb]{0.937,0.937,0.937}}\textbf{\textcolor{red}{0.0121}} & {\cellcolor[rgb]{0.937,0.937,0.937}}\textcolor{blue}{\textbf{0.0374}} & {\cellcolor[rgb]{0.937,0.937,0.937}}\textbf{\textcolor{red}{0.0427}} & {\cellcolor[rgb]{0.937,0.937,0.937}}\textbf{\textcolor{red}{0.2124}} & {\cellcolor[rgb]{0.937,0.937,0.937}}\textbf{\textcolor{red}{0.2387}} & {\cellcolor[rgb]{0.937,0.937,0.937}}\textcolor{red}{\textbf{0.0240}} & {\cellcolor[rgb]{0.937,0.937,0.937}}\textcolor{red}{\textbf{0.0364}} & {\cellcolor[rgb]{0.937,0.937,0.937}}\textcolor{red}{\textbf{0.0155}} & {\cellcolor[rgb]{0.937,0.937,0.937}}\textcolor{red}{\textbf{0.0255}} \\ \hline\hline
\multirow{8}{*}{ArcFace†} & SER-FIQ & 0.0023 & 0.0028 & 0.0069 & 0.0085 & 0.0390 & 0.0439 & \textcolor{blue}{\textbf{0.1947}} & \textbf{\textcolor{red}{0.2347}} & \textcolor{blue}{\textbf{0.0219}} & 0.0330 & 0.0156 & 0.0235 \\
 & SDD-FIQA & \textbf{\textcolor{red}{0.0013}} & \textbf{\textcolor{red}{0.0020}} & 0.0077 & 0.0098 & 0.0468 & 0.0504 & 0.2649 & 0.2930 & 0.0270 & 0.0383 & 0.0185 & 0.0267 \\
 & MagFace & 0.0017 & 0.0022 & 0.0074 & 0.0091 & 0.0495 & 0.0532 & 0.3730 & 0.3996 & 0.0247 & 0.0355 & 0.0171 & 0.0251 \\
 & CR-FIQA(S) & 0.0017 & 0.0023 & 0.0091 & 0.0111 & 0.0409 & 0.0460 & 0.2384 & 0.2757 & 0.0275 & 0.0398 & 0.0185 & 0.0272 \\
 & CR-FIQA(L) & 0.0018 & 0.0024 & \textbf{\textcolor{red}{0.0050}} & \textbf{\textcolor{red}{0.0062}} & \textcolor{blue}{\textbf{0.0371}} & \textcolor{blue}{\textbf{0.0410}} & 0.2055 & 0.2538 & 0.0222 & \textcolor{blue}{\textbf{0.0328}} & \textcolor{blue}{\textbf{0.0149}} & \textcolor{blue}{\textbf{0.0225}} \\
 & IG-FIQA(S-)(Our) & 0.0016 & 0.0022 & 0.0070 & 0.0097 & 0.0396 & 0.0437 & 0.2339 & 0.2688 & 0.0268 & 0.0383 & 0.0177 & 0.0258 \\
 & {\cellcolor[rgb]{0.937,0.937,0.937}}IG-FIQA(S)(Our) & {\cellcolor[rgb]{0.937,0.937,0.937}}0.0017 & {\cellcolor[rgb]{0.937,0.937,0.937}}0.0022 & {\cellcolor[rgb]{0.937,0.937,0.937}}0.0077 & {\cellcolor[rgb]{0.937,0.937,0.937}}0.0101 & {\cellcolor[rgb]{0.937,0.937,0.937}}0.0374 & {\cellcolor[rgb]{0.937,0.937,0.937}}0.0415 & {\cellcolor[rgb]{0.937,0.937,0.937}}0.2059 & {\cellcolor[rgb]{0.937,0.937,0.937}}\textcolor{blue}{\textbf{0.2390}} & {\cellcolor[rgb]{0.937,0.937,0.937}}0.0245 & {\cellcolor[rgb]{0.937,0.937,0.937}}0.0351 & {\cellcolor[rgb]{0.937,0.937,0.937}}0.0166 & {\cellcolor[rgb]{0.937,0.937,0.937}}0.0241 \\
 & {\cellcolor[rgb]{0.937,0.937,0.937}}IG-FIQA(L)(Our) & {\cellcolor[rgb]{0.937,0.937,0.937}}\textcolor{blue}{\textbf{0.0016}} & {\cellcolor[rgb]{0.937,0.937,0.937}}\textcolor{blue}{\textbf{0.0022}} & {\cellcolor[rgb]{0.937,0.937,0.937}}\textcolor{blue}{\textbf{0.0052}} & {\cellcolor[rgb]{0.937,0.937,0.937}}\textcolor{blue}{\textbf{0.0063}} & {\cellcolor[rgb]{0.937,0.937,0.937}}\textcolor{red}{\textbf{0.0371}} & {\cellcolor[rgb]{0.937,0.937,0.937}}\textcolor{red}{\textbf{0.0407}} & {\cellcolor[rgb]{0.937,0.937,0.937}}\textbf{\textcolor{red}{0.1940}} & {\cellcolor[rgb]{0.937,0.937,0.937}}0.2405 & {\cellcolor[rgb]{0.937,0.937,0.937}}\textcolor{red}{\textbf{0.0217}} & {\cellcolor[rgb]{0.937,0.937,0.937}}\textcolor{red}{\textbf{0.0316}} & {\cellcolor[rgb]{0.937,0.937,0.937}}\textcolor{red}{\textbf{0.0146}} & {\cellcolor[rgb]{0.937,0.937,0.937}}\textcolor{red}{\textbf{0.0217}} \\ \hline\hline
\multirow{8}{*}{CurricularFace} & SER-FIQ & 0.0024 & 0.0028 & 0.0092 & 0.0122 & 0.0345 & 0.0574 & \textcolor{blue}{\textbf{0.1664}} & \textbf{\textcolor{red}{0.1969}} & \textcolor{blue}{\textbf{0.0223}} & 0.0332 & 0.0156 & 0.0238 \\
 & SDD-FIQA & \textbf{\textcolor{red}{0.0016}} & \textcolor{blue}{\textbf{0.0022}} & 0.0112 & 0.0144 & 0.0413 & 0.0664 & 0.2320 & 0.2613 & 0.0271 & 0.0389 & 0.0184 & 0.0273 \\
 & MagFace & \textcolor{blue}{\textbf{0.0017}} & \textbf{\textcolor{red}{0.0022}} & 0.0098 & 0.0122 & 0.0448 & 0.0666 & 0.3543 & 0.3966 & 0.0253 & 0.0358 & 0.0174 & 0.0254 \\
 & CR-FIQA(S) & 0.0021 & 0.0027 & 0.0120 & 0.0150 & 0.0350 & 0.0586 & 0.2162 & 0.2585 & 0.0279 & 0.0408 & 0.0185 & 0.0276 \\
 & CR-FIQA(L) & 0.0023 & 0.0029 & \textbf{\textcolor{red}{0.0071}} & \textbf{\textcolor{red}{0.0090}} & \textbf{\textcolor{blue}{0.0330}} & \textcolor{red}{\textbf{0.0507}} & 0.1762 & 0.2579 & 0.0226 & \textcolor{blue}{\textbf{0.0332}} & \textcolor{blue}{\textbf{0.0151}} & \textcolor{blue}{\textbf{0.0230}} \\
 & IG-FIQA(S-)(Our) & 0.0019 & 0.0023 & 0.0112 & 0.0143 & 0.0348 & 0.0596 & 0.2145 & 0.2586 & 0.0269 & 0.0389 & 0.0175 & 0.0262 \\
 & {\cellcolor[rgb]{0.937,0.937,0.937}}IG-FIQA(S)(Our) & {\cellcolor[rgb]{0.937,0.937,0.937}}0.0018 & {\cellcolor[rgb]{0.937,0.937,0.937}}0.0022 & {\cellcolor[rgb]{0.937,0.937,0.937}}0.0100 & {\cellcolor[rgb]{0.937,0.937,0.937}}0.0133 & {\cellcolor[rgb]{0.937,0.937,0.937}}0.0330 & {\cellcolor[rgb]{0.937,0.937,0.937}}0.0540 & {\cellcolor[rgb]{0.937,0.937,0.937}}\textbf{\textcolor{red}{0.1680}} & {\cellcolor[rgb]{0.937,0.937,0.937}}\textcolor{blue}{\textbf{0.2086}} & {\cellcolor[rgb]{0.937,0.937,0.937}}0.0249 & {\cellcolor[rgb]{0.937,0.937,0.937}}0.0360 & {\cellcolor[rgb]{0.937,0.937,0.937}}0.0166 & {\cellcolor[rgb]{0.937,0.937,0.937}}0.0248 \\
 & {\cellcolor[rgb]{0.937,0.937,0.937}}IG-FIQA(L)(Our) & {\cellcolor[rgb]{0.937,0.937,0.937}}0.0017 & {\cellcolor[rgb]{0.937,0.937,0.937}}0.0022 & {\cellcolor[rgb]{0.937,0.937,0.937}}\textcolor{blue}{\textbf{0.0071}} & {\cellcolor[rgb]{0.937,0.937,0.937}}\textcolor{blue}{\textbf{0.0095}} & {\cellcolor[rgb]{0.937,0.937,0.937}}\textcolor{red}{\textbf{0.0329}} & {\cellcolor[rgb]{0.937,0.937,0.937}}\textcolor{blue}{\textbf{0.0532}} & {\cellcolor[rgb]{0.937,0.937,0.937}}0.1692 & {\cellcolor[rgb]{0.937,0.937,0.937}}0.2239 & {\cellcolor[rgb]{0.937,0.937,0.937}}\textcolor{red}{\textbf{0.0222}} & {\cellcolor[rgb]{0.937,0.937,0.937}}\textcolor{red}{\textbf{0.0323}} & {\cellcolor[rgb]{0.937,0.937,0.937}}\textcolor{red}{\textbf{0.0148}} & {\cellcolor[rgb]{0.937,0.937,0.937}}\textcolor{red}{\textbf{0.0224}} \\ \hline\hline
\multirow{8}{*}{MagFace} & SER-FIQ & 0.0024 & 0.0028 & 0.0088 & 0.0094 & 0.0382 & 0.0673 & \textcolor{blue}{\textbf{0.1804}} & 0.2241 & \textcolor{blue}{\textbf{0.0218}} & \textcolor{blue}{\textbf{0.0302}} & 0.0148 & 0.0211 \\
 & SDD-FIQA & \textbf{\textcolor{red}{0.0016}} & 0.0023 & 0.0107 & 0.0122 & 0.0446 & 0.0901 & 0.2414 & 0.2852 & 0.0268 & 0.0372 & 0.0177 & 0.0249 \\
 & MagFace & \textcolor{blue}{\textbf{0.0017}} & \textcolor{blue}{\textbf{0.0023}} & 0.0084 & 0.0097 & 0.0481 & 0.0736 & 0.3552 & 0.4023 & 0.0247 & 0.0343 & 0.0164 & 0.0232 \\
 & CR-FIQA(S) & 0.0020 & 0.0029 & 0.0108 & 0.0142 & 0.0386 & 0.0651 & 0.2175 & 0.2504 & 0.0282 & 0.0388 & 0.0182 & 0.0253 \\
 & CR-FIQA(L) & 0.0022 & 0.0028 & \textbf{\textcolor{red}{0.0055}} & \textbf{\textcolor{red}{0.0069}} & \textcolor{blue}{\textbf{0.0358}} & \textcolor{red}{\textbf{0.0502}} & 0.1852 & 0.2136 & 0.0222 & 0.0309 & \textcolor{blue}{\textbf{0.0144}} & \textcolor{blue}{\textbf{0.0205}} \\
 & IG-FIQA(S-)(Our) & 0.0019 & 0.0026 & 0.0084 & 0.0116 & 0.0378 & 0.0747 & 0.2103 & 0.2280 & 0.0271 & 0.0370 & 0.0173 & 0.0240 \\
 & {\cellcolor[rgb]{0.937,0.937,0.937}}IG-FIQA(S)(Our) & {\cellcolor[rgb]{0.937,0.937,0.937}}0.0018 & {\cellcolor[rgb]{0.937,0.937,0.937}}0.0024 & {\cellcolor[rgb]{0.937,0.937,0.937}}0.0092 & {\cellcolor[rgb]{0.937,0.937,0.937}}0.0112 & {\cellcolor[rgb]{0.937,0.937,0.937}}\textcolor{red}{\textbf{0.0357}} & {\cellcolor[rgb]{0.937,0.937,0.937}}0.0610 & {\cellcolor[rgb]{0.937,0.937,0.937}}0.1827 & {\cellcolor[rgb]{0.937,0.937,0.937}}\textbf{\textcolor{red}{0.2021}} & {\cellcolor[rgb]{0.937,0.937,0.937}}0.0247 & {\cellcolor[rgb]{0.937,0.937,0.937}}0.0336 & {\cellcolor[rgb]{0.937,0.937,0.937}}0.0161 & {\cellcolor[rgb]{0.937,0.937,0.937}}0.0222 \\
 & {\cellcolor[rgb]{0.937,0.937,0.937}}IG-FIQA(L)(Our) & {\cellcolor[rgb]{0.937,0.937,0.937}}0.0017 & {\cellcolor[rgb]{0.937,0.937,0.937}}\textbf{\textcolor{red}{0.0022}} & {\cellcolor[rgb]{0.937,0.937,0.937}}\textcolor{blue}{\textbf{0.0059}} & {\cellcolor[rgb]{0.937,0.937,0.937}}\textcolor{blue}{\textbf{0.0071}} & {\cellcolor[rgb]{0.937,0.937,0.937}}0.0360 & {\cellcolor[rgb]{0.937,0.937,0.937}}\textbf{\textcolor{blue}{0.0609}} & {\cellcolor[rgb]{0.937,0.937,0.937}}\textbf{\textcolor{red}{0.1747}} & {\cellcolor[rgb]{0.937,0.937,0.937}}\textcolor{blue}{\textbf{0.2133}} & {\cellcolor[rgb]{0.937,0.937,0.937}}\textcolor{red}{\textbf{0.0217}} & {\cellcolor[rgb]{0.937,0.937,0.937}}\textcolor{red}{\textbf{0.0298}} & {\cellcolor[rgb]{0.937,0.937,0.937}}\textcolor{red}{\textbf{0.0140}} & {\cellcolor[rgb]{0.937,0.937,0.937}}\textcolor{red}{\textbf{0.0197}} \\ \hline\hline
\multirow{8}{*}{ElasticFace} & SER-FIQ & 0.0021 & 0.0025 & 0.0071 & 0.0132 & 0.0391 & 0.0569 & \textcolor{blue}{\textbf{0.1678}} & 0.2029 & \textcolor{blue}{\textbf{0.0234}} & \textcolor{blue}{\textbf{0.0337}} & 0.0164 & 0.0249 \\
 & SDD-FIQA & \textbf{\textcolor{red}{0.0012}} & \textbf{\textcolor{red}{0.0017}} & 0.0079 & 0.0121 & 0.0446 & 0.0607 & 0.2572 & 0.3127 & 0.0286 & 0.0406 & 0.0196 & 0.0287 \\
 & MagFace & 0.0014 & 0.0019 & 0.0071 & 0.0133 & 0.0483 & 0.0643 & 0.3675 & 0.4249 & 0.0262 & 0.0368 & 0.0182 & 0.0263 \\
 & CR-FIQA(S) & 0.0016 & 0.0021 & 0.0087 & 0.0139 & 0.0391 & 0.0581 & 0.2214 & 0.2901 & 0.0296 & 0.0418 & 0.0199 & 0.0289 \\
 & CR-FIQA(L) & 0.0017 & 0.0022 & \textcolor{blue}{\textbf{0.0056}} & \textcolor{blue}{\textbf{0.0096}} & 0.0358 & \textcolor{blue}{\textbf{0.0517}} & 0.1719 & \textcolor{blue}{\textbf{0.2012}} & 0.0238 & 0.0337 & \textcolor{blue}{\textbf{0.0161}} & \textcolor{blue}{\textbf{0.0235}} \\
 & IG-FIQA(S-)(Our) & 0.0015 & 0.0020 & 0.0069 & 0.0122 & 0.0378 & 0.0551 & 0.2216 & 0.2566 & 0.0286 & 0.0395 & 0.0188 & 0.0271 \\
 & {\cellcolor[rgb]{0.937,0.937,0.937}}IG-FIQA(S)(Our) & {\cellcolor[rgb]{0.937,0.937,0.937}}0.0015 & {\cellcolor[rgb]{0.937,0.937,0.937}}0.0020 & {\cellcolor[rgb]{0.937,0.937,0.937}}0.0066 & {\cellcolor[rgb]{0.937,0.937,0.937}}0.0121 & {\cellcolor[rgb]{0.937,0.937,0.937}}\textcolor{blue}{\textbf{0.0356}} & {\cellcolor[rgb]{0.937,0.937,0.937}}0.0524 & {\cellcolor[rgb]{0.937,0.937,0.937}}0.1772 & {\cellcolor[rgb]{0.937,0.937,0.937}}0.2033 & {\cellcolor[rgb]{0.937,0.937,0.937}}0.0266 & {\cellcolor[rgb]{0.937,0.937,0.937}}0.0366 & {\cellcolor[rgb]{0.937,0.937,0.937}}0.0179 & {\cellcolor[rgb]{0.937,0.937,0.937}}0.0255 \\
 & {\cellcolor[rgb]{0.937,0.937,0.937}}IG-FIQA(L)(Our) & {\cellcolor[rgb]{0.937,0.937,0.937}}\textcolor{blue}{\textbf{0.0013}} & {\cellcolor[rgb]{0.937,0.937,0.937}}\textcolor{blue}{\textbf{0.0018}} & {\cellcolor[rgb]{0.937,0.937,0.937}}\textbf{\textcolor{red}{0.0053}} & {\cellcolor[rgb]{0.937,0.937,0.937}}\textbf{\textcolor{red}{0.0094}} & {\cellcolor[rgb]{0.937,0.937,0.937}}\textcolor{red}{\textbf{0.0355}} & {\cellcolor[rgb]{0.937,0.937,0.937}}\textcolor{red}{\textbf{0.0512}} & {\cellcolor[rgb]{0.937,0.937,0.937}}\textbf{\textcolor{red}{0.1631}} & {\cellcolor[rgb]{0.937,0.937,0.937}}\textbf{\textcolor{red}{0.1908}} & {\cellcolor[rgb]{0.937,0.937,0.937}}\textcolor{red}{\textbf{0.0234}} & {\cellcolor[rgb]{0.937,0.937,0.937}}\textcolor{red}{\textbf{0.0329}} & {\cellcolor[rgb]{0.937,0.937,0.937}}\textcolor{red}{\textbf{0.0158}} & {\cellcolor[rgb]{0.937,0.937,0.937}}\textcolor{red}{\textbf{0.0231}} \\ \hline\hline
\multirow{8}{*}{AdaFace} & SER-FIQ & 0.0024 & 0.0028 & 0.0081 & 0.0129 & 0.0329 & 0.0389 & \textcolor{blue}{\textbf{0.1312}} & \textcolor{blue}{\textbf{0.1785}} & \textcolor{blue}{\textbf{0.0181}} & \textcolor{blue}{\textbf{0.0257}} & 0.0121 & 0.0176 \\
 & SDD-FIQA & \textbf{\textcolor{red}{0.0016}} & 0.0022 & 0.0096 & 0.0162 & 0.0409 & 0.0469 & 0.1911 & 0.2646 & 0.0220 & 0.0313 & 0.0147 & 0.0210 \\
 & MagFace & \textcolor{blue}{\textbf{0.0017}} & \textbf{\textcolor{red}{0.0022}} & 0.0079 & 0.0125 & 0.0443 & 0.0498 & 0.3107 & 0.3352 & 0.0205 & 0.0289 & 0.0136 & 0.0193 \\
 & CR-FIQA(S) & 0.0021 & 0.0026 & 0.0107 & 0.0171 & 0.0347 & 0.0410 & 0.1798 & 0.2188 & 0.0229 & 0.0322 & 0.0146 & 0.0208 \\
 & CR-FIQA(L) & 0.0023 & 0.0028 & \textbf{\textcolor{red}{0.0062}} & \textcolor{blue}{\textbf{0.0097}} & \textcolor{red}{\textbf{0.0322}} & \textcolor{blue}{\textbf{0.0377}} & 0.1542 & 0.2126 & 0.0183 & 0.0262 & \textcolor{blue}{\textbf{0.0119}} & \textcolor{blue}{\textbf{0.0171}} \\
 & IG-FIQA(S-)(Our) & 0.0019 & 0.0023 & 0.0091 & 0.0146 & 0.0341 & 0.0400 & 0.1708 & 0.2107 & 0.0225 & 0.0313 & 0.0143 & 0.0200 \\
 & {\cellcolor[rgb]{0.937,0.937,0.937}}IG-FIQA(S)(Our) & {\cellcolor[rgb]{0.937,0.937,0.937}}0.0018 & {\cellcolor[rgb]{0.937,0.937,0.937}}0.0022 & {\cellcolor[rgb]{0.937,0.937,0.937}}0.0098 & {\cellcolor[rgb]{0.937,0.937,0.937}}0.0164 & {\cellcolor[rgb]{0.937,0.937,0.937}}0.0323 & {\cellcolor[rgb]{0.937,0.937,0.937}}0.0379 & {\cellcolor[rgb]{0.937,0.937,0.937}}0.1432 & {\cellcolor[rgb]{0.937,0.937,0.937}}0.1890 & {\cellcolor[rgb]{0.937,0.937,0.937}}0.0205 & {\cellcolor[rgb]{0.937,0.937,0.937}}0.0283 & {\cellcolor[rgb]{0.937,0.937,0.937}}0.0132 & {\cellcolor[rgb]{0.937,0.937,0.937}}0.0186 \\
 & {\cellcolor[rgb]{0.937,0.937,0.937}}IG-FIQA(L)(Our) & {\cellcolor[rgb]{0.937,0.937,0.937}}0.0017 & {\cellcolor[rgb]{0.937,0.937,0.937}}\textcolor{blue}{\textbf{0.0022}} & {\cellcolor[rgb]{0.937,0.937,0.937}}\textcolor{blue}{\textbf{0.0062}} & {\cellcolor[rgb]{0.937,0.937,0.937}}\textbf{\textcolor{red}{0.0096}} & {\cellcolor[rgb]{0.937,0.937,0.937}}\textcolor{blue}{\textbf{0.0322}} & {\cellcolor[rgb]{0.937,0.937,0.937}}\textcolor{red}{\textbf{0.0377}} & {\cellcolor[rgb]{0.937,0.937,0.937}}\textbf{\textcolor{red}{0.1268}} & {\cellcolor[rgb]{0.937,0.937,0.937}}\textbf{\textcolor{red}{0.1712}} & {\cellcolor[rgb]{0.937,0.937,0.937}}\textcolor{red}{\textbf{0.0180}} & {\cellcolor[rgb]{0.937,0.937,0.937}}\textcolor{red}{\textbf{0.0254}} & {\cellcolor[rgb]{0.937,0.937,0.937}}\textcolor{red}{\textbf{0.0116}} & {\cellcolor[rgb]{0.937,0.937,0.937}}\textcolor{red}{\textbf{0.0165}} \\ \hline
\end{tabular}
}
\caption{AUCs of ERC obtained by recent SOTA FIQA methods and suggesting IG-FIQA. Annotated † means re-implemented, and without annotation means used pretrained model provided from official repository. IG-FIQA(S-) refers to the IG-FIQA(S) without augmentation. \color{red}\textbf{Red} \color{black}: best, \color{blue}\textbf{Blue} \color{black}: second.}
\label{bench_table}
\vspace*{-5mm}
\end{table*}
\subsection{Comparison with SOTA methods}
We compared the FIQA performance of the proposed model, IG-FIQA, with the recently presented SOTA models: SER-FIQ \cite{terhorst2020ser}, SDD-FIQA \cite{ou2021sdd}, MagFace \cite{meng2021magface}, and CR-FIQA \cite{boutros2023cr}. In Tab. \ref{bench_table}, we annotated the AUCs as verification performance at FMR=1e-3 and FMR=1e-4. ERCs using ArcFace and AdaFace are reported in Fig. \ref{erc_plot1} and Fig. \ref{erc_plot2}. For a more detailed analysis of the impact of data augmentation, we also include a small protocol model trained without data augmentation (IG-FIQA(S-)) in the results.

From Fig. \ref{erc_plot1} and Fig. \ref{erc_plot2}, we can see that all FIQA methods fluctuate on the LFW benchmark and can not infer quality properly. This is because current SOTA FR models have already reached saturation in the LFW benchmark; in other words, FR models can extract feature robustly while ignoring minor quality degradation. Since IG-FIQA uses data augmentation to generate images of various qualities and uses them for training, there is a risk of poor performance on high-quality benchmarks compared to other FIQA models trained only on high-quality datasets. In fact, it can be seen that IG-FIQA(S-) performs better than IG-FIQA(S) in CFP-FP benchmark. Nevertheless, the proposed IG-FIQA achieved similar or slightly better performance than the conventional SOTA methods in the CPLFW and CFP-FP benchmarks.

From a FIQA perspective, the original purpose of FIQA is to select good quality facial images from multiple mixed-quality images to ensure reliable FR algorithm performance. However, with the emergence of high-performance FR models, FR performance on high-quality datasets has become saturated. Therefore, FIQA is less necessary for high-quality datasets, and it can be difficult to distinguish between superior FIQA methods. Noteworthily, IG-FIQA outperforms most SOTA models on the mixed-quality benchmark datasets. This indicates that the proposed IG-FIQA is an effective FIQA model capable of filtering low-quality images from images of varying qualities, aligning with the original purpose of FIQA. Verification pairs for the XQLFW benchmark are selected based on SER-FIQ and BRISQUE \cite{mittal2012no} scores, which may give an advantage to SER-FIQ. Nevertheless, IG-FIQA outperforms SER-FIQ on XQLFW and achieves SOTA. Additionally, the performance gap between small and large protocols of the proposed method is much smaller on various benchmarks than that of CR-FIQA. This means that IG-FIQA is capable of generalizing the regression network effectively, even with small training datasets and a lightweight FR backbone. Our small protocol model without augmentation (IG-FIQA(S-)) consistently exhibits better performance than CR-FIQA(S). This proves that the proposed method of removing classes that are at risk of being mislabeled during training helps improve performance.

\section{Conclusions}
In this paper, we address the limitations of the conventional SOTA FIQA method that use sample relative classifiability as pseudo-labels. This approach often assign inaccurate pseudo-labels to images with low intra-class variation, regardless of their actual quality. The proposed novel method is simple yet very effective in identifying classes that are at risk of being mislabeled during training and excluding them from the training process, incurring negligible computational cost. Our method does not require a pre-processing for data cleaning or a pre-trained model and can be trained in an end-to-end manner. Additionally, by introducing a pipeline that can safely apply data augmentation in sample relative classifiability method, our proposed approach outperforms existing methods across various benchmarks, thereby establishing a new SOTA in the field of FIQA.

\par\vfill\par

%
%
\bibliographystyle{splncs04}
\bibliography{main}
\end{document}